\documentclass[10pt, a4paper]{article}

\usepackage{lrec-coling2024} 

\usepackage{graphicx}
\usepackage{tabularx}
\usepackage{soul}
\usepackage{subfig}
\usepackage{amsmath}
\usepackage{amssymb}
\usepackage{multirow,makecell}
\usepackage{float}

\title{Multi-modal Semantic Understanding 
\\ with Contrastive Cross-modal Feature Alignment}

\name{Ming Zhang$^{1, 2\ast}$, Ke Chang$^{1, 3\ast}$, Yunfang Wu$^{1, 3\dagger}$}
\address{$^1$National Key Laboratory for Multimedia Information Processing, Peking University\\
         $^2$School of Software and Microelectronics, Peking University \\
         $^3$School of Computer Science, Peking University\\         zhangming@stu.pku.edu.cn, \{changkegg, wuyf\}@pku.edu.cn\\}

\abstract{
Multi-modal semantic understanding requires integrating information from different modalities to extract users' real intention behind words. Most previous work applies a dual-encoder structure to separately encode image and text, but fails to learn cross-modal feature alignment, making it hard to achieve cross-modal deep information interaction. This paper proposes a novel CLIP-guided contrastive-learning-based architecture to perform multi-modal feature alignment, which projects the features derived from different modalities into a unified deep space. On multi-modal sarcasm detection (MMSD) and multi-modal sentiment analysis (MMSA) tasks, the experimental results show that our proposed model significantly outperforms  
several baselines, 
and our feature alignment strategy brings obvious performance gain over models with different aggregating methods and models even enriched with knowledge. 
More importantly, our model is simple to implement without using task-specific external knowledge, and thus can easily migrate to other multi-modal tasks. Our source codes are available at \href{https://github.com/ChangKe123/CLFA}{https://github.com/ChangKe123/CLFA}.
\\ \newline  }

\begin{document}

\maketitleabstract

\section{Introduction}
\renewcommand{\thefootnote}{\fnsymbol{footnote}}
\footnotetext[1]{Equal contribution.}
\renewcommand{\thefootnote}{\arabic{footnote}}

Recently, with the increasing use of various forms of media such as text, image, video, and audio for information communication on social platforms, semantic understanding involving multi-modal data becomes urgently needed, where sentiment analysis and sarcasm detection are two important tasks.

The multi-modal sentiment analysis task (MMSA) takes data in more than one modal as input, and outputs a classification result with positive, negative and neutral. 

\begin{figure}[ht]
  \centering
  \subfloat[Some good stuff right here.]{
    \label{sfig:pos}
    \includegraphics[width=0.225\textwidth]{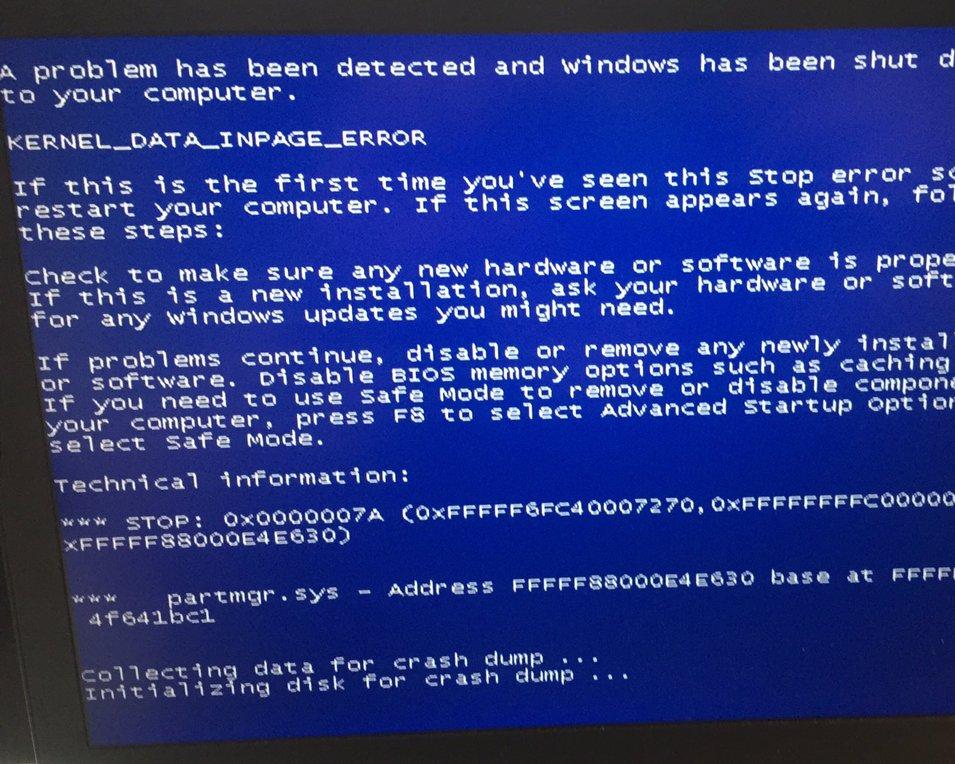}}
    \hfill
   \subfloat[Fitness whole pizza in my mouth!]{
    \label{sfig:neg}
    \includegraphics[width=0.225\textwidth]{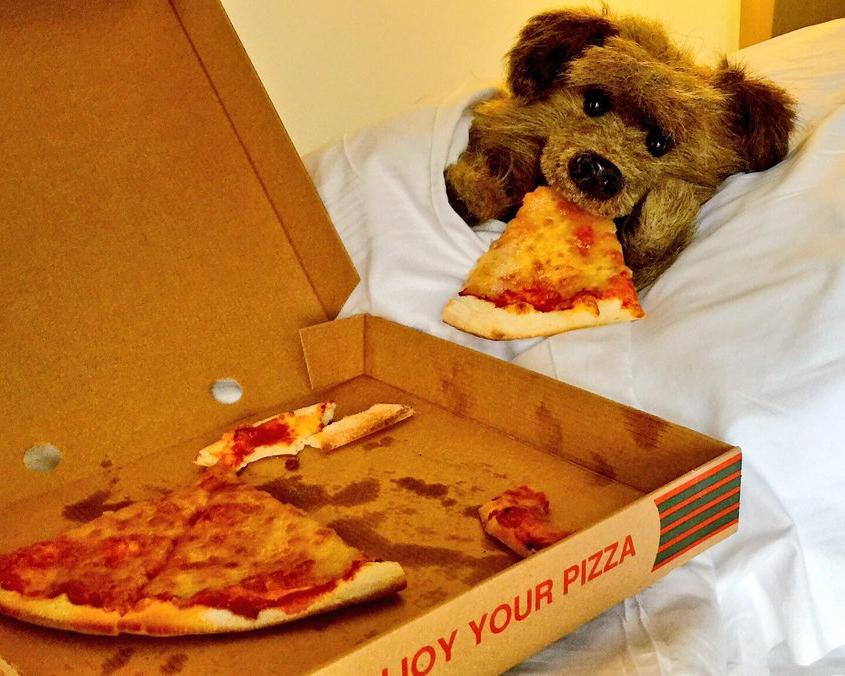}}
  \caption{Multi-modal sarcastic examples. The left text-image pair conveys sarcasm, while the right has no sarcasm.}
  \label{fig:example}
\end{figure}

The multi-modal sarcasm detection task (MMSD) is more complex and difficult. People often use sarcasm to humorously express mockery, ridicule, criticism, and other emotions towards certain individuals or events. Therefore, sarcasm detection aids in uncovering users' real intentions behind the words, thereby enhancing the effectiveness of semantic understanding. For instance in Figure \ref{fig:example}, the sarcasm example contains a twitter "some good stuff here" paired with a crash computer picture. It is likely to misinterpret the intended opinion from the text alone, leading to a biased result. 

Current research on multi-modal semantic understanding primarily exploits a dual-encoder structure, which utilizes two separate encoders to process image and text. The extracted features from different modalities are then fused to capture cross-modal information. Previous methods for feature fusion include simple concatenation ~\cite{schifanella2016detecting}, attention mechanism~\cite{xu-etal-2020-reasoning,pan-etal-2020-modeling} and 
graph neural networks~\cite{liang2021multi,liu-etal-2022-towards-multi-modal,liang-etal-2022-multi}.
However, these methods 
neglect to consider the semantic alignment between different modalities when fusing multi-modal features. The features 
derived from image and text data and encoded by separate encoders exist in different semantic spaces. Consequently, utilizing attention or graph neural networks to fuse multi-modal features directly leads to a semantic mismatch, making it hard to learn an effective cross-modal interaction. 

In this paper, we propose a novel semantic alignment method for multi-modal 
deep understanding tasks, 
namely CLIP-guided 
\textbf{C}ontrastive-\textbf{L}earning-based \textbf{F}eature \textbf{A}lignment (CLFA). Specially, we leverage BERT~\cite{devlin-etal-2019-bert} to encode text and ViT~\cite{dosovitskiy2020image} to encode image. We borrow CLIP~\cite{radford2021learning} as the teacher model and employ contrastive learning to achieve semantic alignment between text and image, 
projecting the representations derived from different modalities and different encoders into a unified space. 
We devise a multi-task learning architecture, by treating the feature alignment as 
an auxiliary task to facilitate the main classification task. 
In this way, our proposed method  
can utilize the powerful ability of CLIP and 
modal-related encoding simultaneously. 

We conduct extensive experiments on the public MMSA dataset~\cite{MVSA} and MMSD dataset~\cite{cai-etal-2019-multi}. Our proposed model CLFA brings obvious performance gains over baselines, achieving competitive results with models incorporating external knowledge, 
including word sentiment, dependency tree and image captions. Besides, we integrate CLFA into a knowledge-enhanced model and several multi-modal fusion methods, achieving better performance and verifying the effectiveness and versatility of our model.    
More importantly, without using external task-related knowledge and third-party tools, our method can be easily adapted to other multi-modal tasks. 

To sum up, our contributions are as follows:

  \begin{itemize}
    \item{We propose a novel CLIP-guided contrastive-learning-based architecture for multi-modal semantic alignment, which injects different sorts of features derived from heterogeneous text and image modalities into a unified space.}
    \item{Our simple method significantly outperforms 
    baseline models and achieves comparable results with knowledge-enhanced models, on two multi-modal semantic understanding tasks including sentiment analysis and sarcasm detection.}
    \item{Plenty of experiments show that our proposed method is effective on different cross-modal aggregating methods. 
    Besides, our method can be combined with other knowledge-based models to get higher performance.}
 \end{itemize}

 \begin{figure*}[t]
  \centering
      \includegraphics[width=0.8\linewidth]{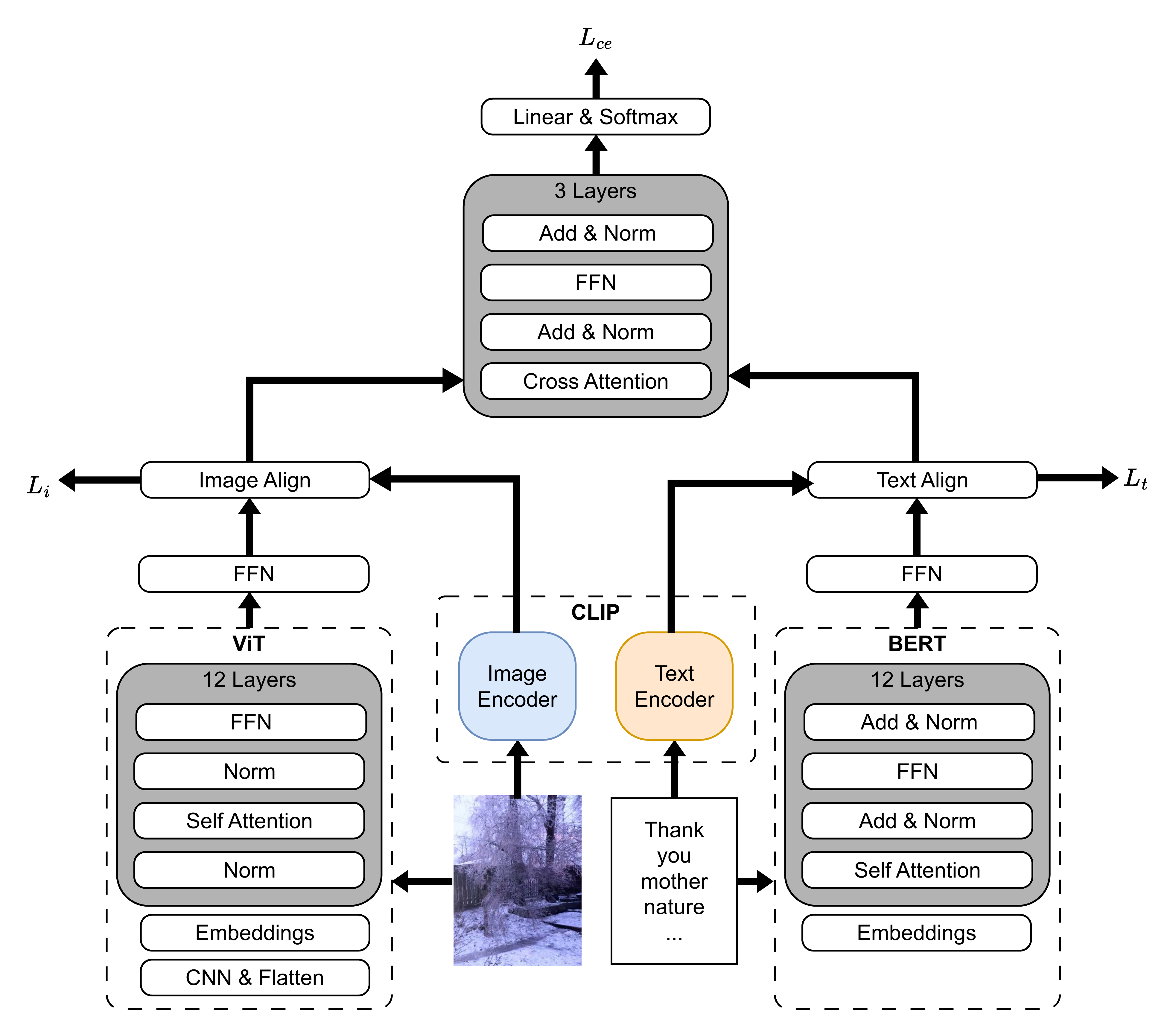}
  \caption{Overview of our proposed model.}
  \label{fig:model}
\end{figure*}

\section{Methodology}
The overall structure of our proposed model CLFA is illustrated in Figure \ref{fig:model}. 
Specially, the text and image inputs are separately fed into BERT~\cite{devlin-etal-2019-bert} and ViT~\cite{dosovitskiy2020image} to obtain their respective representations. 
At the core of our model, different representations from text and image are projected into the same deep space 
guided by the common CLIP representations via contrastive learning, which serves as a subtask to facilitate the main classification task. 
After that, the text and image features are input into a cross attention layer to perform multi-modal fusion, and then a classification layer to predict the result.

\subsection{Text and Image 
Encoding}

Given an input sentence $S = [s_1,s_2,\cdots, s_n]$, where $n$ represents the length, we adopt BERT to 
encode each token 
to the textual representation:
\begin{equation}
\label{e1}
    f_t(S)=\mathbf{BERT}(S)
\end{equation}
The last hidden state $f_t(S) = [h_1, h_2, \cdots, h_n] \in \mathcal{R}^{n \times d}$ 
is considered as the representation of text, 

Following previous studies, we employ the pre-trained model ViT for the image modality. The input image is resized to $224 \times 224$ and divided into $m = 7 \times 7$ image patches, and then flattened into a sequence $P = [P_1,P_2,\cdots, P_m] \in \mathcal{R}^{m \times 32 \times 32}$. Finally, the image 
representation $f_i \in \mathcal{R}^{m \times d}$ is obtained through the encoder of ViT: 
\begin{equation}
\label{e2}
    f_i(P)=\mathbf{ViT}(P)
\end{equation}

We 
adopt BERT and ViT of the same scale, so the parameter $d$ is unified in our experiment.

\subsection{Cross-modal Feature Alignment}
As the text and image are encoded by two independent encoders, the distributions of two types of features are inconsistent. As a result, it is difficult to model the interaction between different modalities without alignment. 
Therefore, we introduce CLIP~\cite{radford2021learning} and employ contrastive learning to enhance the model ability to align multi-modal features.


CLIP is pre-trained on a large-scale dataset of image-text pairs, with the aim of feature aligning. In our dual-encoder structure, CLIP innovatively serves as a teacher model to guide the alignment of text and image modalities through contrastive learning.  

Our contrastive learning is done in each mini-batch, which contains $B$ text-image pairs, $[(T_1, I_1), (T_2, I_2), \cdots, (T_B, I_B)]$. Using CLIP, we obtain the text representation ${C_t}$ and image representation ${C_i}$, respectively:

\begin{equation}
\label{e31}
\begin{split}
    C_t &= [\mathbf{CLIP_t}(T_1), \cdots, \mathbf{CLIP_t}(T_B)]\\
    C_i &= [\mathbf{CLIP_i}(I_1), \cdots, \mathbf{CLIP_i}(I_B)]\\
\end{split}
\end{equation}

As CLIP has different output dimension $d_C$ with BERT and ViT, we add a MLP mapping net, to align the output features of BERT and ViT to the same dimension size of CLIP:

\begin{equation}
\label{e32}
\begin{split}
    f'_t &= \mathbf{MLP}(f_t)\\
    F'_t &= [f'_t(T_1), \cdots, f'_t(T_B)]\\
    f'_i &= \mathbf{MLP}(f_i)\\
    F'_i &= [f'_i(I_1), \cdots, f'_i(I_B)]\\
\end{split}
\end{equation}
That is, the MLP layer takes a $d$-dim vector as input and outputs a $d_C$-dim vector.

\paragraph{Image CLIP-ViT representation alignment.}
For the image alignment, we try to align the CLIP image representation $C_i$ and ViT representation $F'_i$.
Consequently, we consider the matched representations of CLIP and ViT in one mini-batch as positive examples, while unmatched representations as negative examples.  
Two distinct loss functions with contrastive learning are 
constructed:
\begin{itemize}

\item Using the ViT representation as the anchor, denoted as $\mathcal{L}_{ic}$;

\item Using the CLIP representation as the anchor, denoted as $\mathcal{L}_{ci}$.
\end{itemize}

The loss functions are computed
as below:
\begin{equation}
\label{e35}
\begin{split}
    \mathcal{L}_{ic}&=-\frac{1}{B}\sum_{k=1}^B{\log\frac{\mathbf{e}^{\mathbf{sim}(F'_{ik},C_{ik})/\tau}}{\sum_{j=1}^B{\mathbf{e}^{\mathbf{sim}(F'_{ik},C_{ij})/\tau}}}}\\
    \mathcal{L}_{ci}&=-\frac{1}{B}\sum_{k=1}^B{\log\frac{\mathbf{e}^{\mathbf{sim}(C_{ik},F'_{ik})/\tau}}{\sum_{j=1}^B{\mathbf{e}^{\mathbf{sim}(C_{ik},F'_{ij})/\tau}}}}\\
\end{split}
\end{equation}
Here the operation
\textbf{sim} is to calculate the cosine similarity.

Further, the average of these two loss functions is regarded as the 
contrastive learning loss function for the image alignment, denoted as $\mathcal{L}_{i}$:
\begin{equation}
\label{e36}
\begin{split}
    \mathcal{L}_{i}&=\frac{1}{2}\mathcal{L}_{ci}+\frac{1}{2}\mathcal{L}_{ic}
\end{split}
\end{equation}

\paragraph{Text CLIP-BERT representation alignment.}
For the text alignment, we also obtain the contrastive learning loss $\mathcal{L}_{t}$ by aligning the CLIP text representation $C_t$ and BERT representation $F'_t$, in the same way as the image alignment.

\paragraph{Image-Text feature alignment}
With CLIP as guidance, we perform image-text alignment 
by combing the loss functions of image alignment and text alignment. 
The complete contrastive loss function $\mathcal{L}_{con}$ is computed by: 

\begin{equation}
\label{e4}
    \mathcal{L}_{con}=\frac{1}{2}\mathcal{L}_{i}+\frac{1}{2}\mathcal{L}_{t}
\end{equation}

Through this indirect approach, the dual encoders in our model achieve multi-modal feature alignment. 
It is important to notice that our method is not equivalent to using contrastive learning directly to close the gap between image-text pairs, as in the sarcasm detection task, the twitter text and the image are not always semantically consistent. 
Actually, the role of the CLIP teacher model is to project features of individual modalities into a unified space and make multi-modal semantics close.

\subsection{Cross Attention and Classification}

Getting the aligned representations of image and text, we can aggregate them to understand the cross-modal semantics better. For example, we can simply concatenate them and treat the result as a unimodal data for classification. However, in semantic analysis task, the relationship between the two modalities is important and difficult to capture. So we adopt cross attention to discover the complex relationship between text and image, especially finding the inconsistency for the sarcasm detection task.

The main idea in cross attention is using different input to calculate $Q$ and $K, V$. For multi-modal semantic understanding, we regard the text representation as Query and image representation as Key and Value:

\begin{equation}
\begin{split}
    Attn(Q,K,V)&=\mathrm{softmax}(\frac{QK^T}{\sqrt{d_k}})V
\end{split}
\end{equation}

In the next layer, the $F'_t$ is replaced by the hidden state  $h$ of the previous layer: 
\begin{equation}
\begin{split}
    h&=Attn(W_qF'_t, W_kF'_i, W_vF'_i)\\
\end{split}
\end{equation}

After 3 layers 
with cross attention, we get the representation and input it into the classification layers, including FFN and softmax.

The final objective function is the weighted sum of the cross-entropy loss and the contrastive learning loss:
\begin{equation}
\label{e47}
\mathcal{L}=\alpha\mathcal{L}_{con}+\mathcal{L}_{ce}
\end{equation}
where $\alpha$ is a hyper-parameter to balance two parts. 

\section{Experimental Setup}

\subsection{Dataset and Evaluation Metrics}
The publicly available dataset for MMSD was released by~\citet{cai-etal-2019-multi}, which consists of Twitter image-text pairs. 
The training data collected twitters containing hashtags such as "\texttt{\#sarcasm}" as positive samples, and the labels of the validation set and test set were manually verified. The statistical distribution of MMSD dataset is reported in Table 1.

The MVSA dataset ~\cite{MVSA} is a commonly used dataset for the MMSA task. It consists of two parts: Single and Multiple. The Single part is annotated by only one person, and the Multiple part has 3 annotations whose voting results are regarded as golden references by most previous work. The detailed information of MVSA-Single and MVSA-Multipe datasets is listed in Table 2 and Table 3.

The evaluation metrics include accuracy (Acc), precision (P), recall (R), and F1. Following previous researches, we also calculate and report the macro precision, recall and F1. For comparing model performance, we adopt F1 as the main metric.
\begin{table}[h]
    \centering
    \label{tab:t30}
        \label{tab:t31}
        \begin{tabular}{@{}cccccc@{}}
         \Xhline{1pt}
         & Samples & \ Positive & \ Negative  \\
         \Xhline{0.5pt}
            Train  & 19816 & 8642 & 11174  \\
            Dev & 2410 & 959 & 1451  \\
            Test & 2409 & 959 & 1450  \\
         \Xhline{0.5pt}
            Total & 24635 & 10560 & 14075  \\
          \Xhline{1pt}
        \end{tabular}
        \caption[]{Dataset statistics of sarcasm detection~\cite{cai-etal-2019-multi}.}
\end{table}

\begin{table}[h]
    \centering
        \begin{tabular}{@{}cccccc@{}}
         \Xhline{1pt}
         & \ Samples & \ Pos. & Neutral & Neg. \\
         \Xhline{0.5pt}
            Train  & 3611 & 2147 & 376 & 1088  \\
            Dev & 450 & 268 & 47 & 135  \\
            Test & 450 & 268 & 47 & 135  \\
         \Xhline{0.5pt}
            Total & 4511 & 2683 & 470 & 1358  \\
          \Xhline{1pt}
        \end{tabular}
        \caption[]{Dataset statistics of MVSA-Single~\cite{MVSA}.}
\end{table}

\begin{table}[h]
    \centering
        \begin{tabular}{@{}cccccc@{}}
         \Xhline{1pt}
         & Samples & \ Pos. & \ Neutral & \ Neg.  \\
         \Xhline{0.5pt}
            Train  & 13624 & 9056 & 3528 & 1040  \\
            Dev & 1700 & 1131 & 440 & 129  \\
            Test & 1700 & 1131 & 440 & 129  \\
         \Xhline{0.5pt}
            Total & 17024 & 11318 & 4408 & 1298  \\
          \Xhline{1pt}
        \end{tabular}
        \caption{Dataset statistics of MVSA-Multiple~\cite{MVSA}.}
\end{table}

\subsection{Experimental Details}

To encode the text, we use bert-base-uncased from HuggingFace.
We use Google's ViT-B\_32 pretrained model with the image patch size of 32x32 
to obtain the image representation. 
For feature alignment,  
we utilize the clip-vit-base-patch32 pre-trained model. 
The maximum input text length is 77. The Transformer layer with cross-attention is set to 3. 
The mapping net adopts a two-layer perception network with a hidden dimension of 1536 and an output dimension of 512.
During the training, the batch size is set to 8, learning rate is 1e-5, dropout rate is 0.1. The model is trained for 15 epochs. The model employs a warmup strategy with a 
proportion of 0.1. The model parameter's L2 regularization coefficient is 0.01. The contrastive learning temperature coefficient $\tau$ is 0.1.

\subsection{Baseline Methods}

Since the MMSD task is quite harder, we mainly compare our model with 
previous methods on this task. As baselines, we do experiments with the image-based models Resnet~\cite{he2016deep} and ViT~\cite{dosovitskiy2020image}, as well as text-based models BiLSTM and BERT. Additionally, we compare our model with the following multi-modal approaches.

\textbf{CLIP:}~\cite{radford2021learning} A multi-modal pretraining model that aligns text and image features through contrastive learning. The 
aligned text-image features are then concatenated to obtain multi-modal representations. 
We also experiment a model 
by combining CLIP and the cross-modal attention, namely \textbf{CLIP+Cross}.

\textbf{MLP+CNN:}~\cite{schifanella2016detecting} It uses MLP to encode text and CNN for image, and 
the two modalities' features are concatenated and fed into a classification layer for prediction.

\textbf{HFM:}~\cite{cai-etal-2019-multi} 
A hierarchical fusion model that treats image, text, and image attributes as three modalities. 
The multi-modal features are summed according to attention weights to obtain the final representation.

\textbf{D\&R Net:}~\cite{xu-etal-2020-reasoning}
The Decomposed and Relationship Network decomposes image and text features into HFM same space through a shared layer,   
and constructs relationships between text and adjective-noun pairs 
derived from images using attention mechanisms.

\textbf{ResBert:}~\cite{pan-etal-2020-modeling} This model utilizes the pretrained ResNet and BERT models. It extracts hashtags from the input text, and fuses text and image modality features using a cross-modal attention mechanism. 

\textbf{InCross:}~\cite{liang2021multi} By considering the dependency relationships within text and the adjacency relationships within image blocks, this model 
constructs relationship graphs within and between modalities, and performs multi-modal 
feature fusion with graph convolutional neural networks.

\textbf{HKE:}~\cite{liu-etal-2022-towards-multi-modal} A knowledge-enhanced hierarchical model that uses image captions as external knowledge. The model 
captures both atomic-level and compositional-level sentiment incongruity, with attention mechanisms at the atomic level and a graph attention network at the compositional level.

\textbf{CMGCN:}~\cite{liang-etal-2022-multi} This model 
utilizes an image object recognition tool to identify objects and their attributes. 
It constructs cross-modal graph relationships based on image object information, and utilizes graph convolutional neural networks and attention mechanisms for feature fusion.

In the recent models, 
InCross, HKE and CMGCN introduce external knowledge to 
boost performance. 
For example, 
they take the dependency tree as input in the text modality, 
and leverage generated captions or the object detection result for image.
Our model doesn't leverage any external knowledge, and can be combined with these models. 

For MMSA task, we compare our method with the baseline model BERT+ViT, which uses BERT to encode text and ViT to encode image, and then concatenates them for classification. Besides, we use RoBERTa pre-trained on Twitter corpus\footnote{\href{https://huggingface.co/cardiffnlp/twitter-roberta-base}{https://huggingface.co/cardiffnlp/twitter-roberta-base}.} as the text encoder, which is more fit with the MVSA data collected also from the Twitter corpus.  

\section{Results and Analysis}
\subsection{Main Results}
The overall experimental results are reported in Table \ref{tab:t32} and Table \ref{tab:t43}. 

\begin{table*}[htbp]
   \centering
   \renewcommand\arraystretch{1.2}
   \setlength\tabcolsep{7pt}
   \begin{small}
   \begin{tabular}{@{}l|l|ccccccc@{}}
   \Xhline{1pt}
   \multirow{2}{*}{\textbf{Categories}} & \multirow{2}{*}{\textbf{Models}} &  \multirow{2}{*}{\textbf{Acc}(\%)} & \multirow{2}{*}{\textbf{P}(\%)} & \multirow{2}{*}{\textbf{R}(\%)} & \multirow{2}{*}{\textbf{F1}(\%)} & \multicolumn{3}{c}{\textbf{Macro}}  \\ \Xcline{7-9}{0.5pt} 
    & & & & & & \textbf{P}(\%) & \textbf{R}(\%) &  \textbf{F1} (\%)  \\ \Xhline{0.5pt}
    \multirow{2}{*}{Image Based} & Resnet* 
    & 71.27 & 63.02 & 67.36 & 65.12 & 70.20 & 70.61 & 70.35 \\
    &ViT* & 72.15 & 64.72 & 66.01 & 65.36 & 70.97 & 71.11 & 71.03 \\
    \Xhline{0.5pt}
    \multirow{2}{*}{Text Based} & BiLSTM* & 76.21 & 71.59 & 66.74 & 69.08 & 75.27 & 74.61 & 74.88 \\
    &BERT* & 79.95 & 72.2 & 80.71 & 76.22 & 79.18 & 80.08 & 79.44 \\
    \Xhline{0.5pt}
    \multirow{8}{*}{Multi-modal} & CLIP*  & 84.56 & \textbf{84.57}&74.87 &79.42 &	84.56 &	82.92 &	83.53  \\
    &CLIP+Cross Attention* &  85.14 & 80.82&	82.17 &81.49 &	84.45 &	84.64 &	84.54 \\
    &MLP+CNN & 81.61 & - & - & - & 79.52 & 72.47 & 75.83 \\
    &HFM &  83.44 & 76.57 & 84.15 & 80.18 & 79.40 & 82.45 & 80.90 \\
    &D\&R Net  & 84.02 & 77.97 & 83.42 & 80.60 & - & - & - \\
    &ResBert & 86.05 & 78.63 & 83.31 & 80.90 & 78.87 & 84.46 & 82.92 
    \\     \Xcline{2-9}{0.5pt}
    & BERT+ViT* & 83.73 & 78.12 & 81.54  & 79.80 & 82.94 & 83.41 & 83.15\\
    & \textbf{Our CLFA}* &  \textbf{86.80} & 81.51 & \textbf{86.44} & \textbf{83.91} & \textbf{86.09} & \textbf{86.74} & \textbf{86.36}  \\
     \hline
     \hline
     \multirow{3}{*}{With Knowledge} & InCross &  86.10 & 81.38 & 84.36 & 82.84 & 85.39 & 85.80 & 85.60 \\
    &HKE & 87.36 & 81.84 & \textbf{86.48} & 84.09 & - & - & - \\
    &CMGCN & \textbf{87.55} & \textbf{83.63} & 84.69 & \textbf{84.16} & \textbf{87.02} & \textbf{86.97} & \textbf{87.00} \\ 
    \Xhline{1pt}
   \end{tabular}\\[6pt]
   \end{small}

   \caption{Experimental results on MMSD. Those marked with * represent the results of our experiments, while others are retrieved from the published paper. The best results among each group (without knowledge and with knowledge) are highlighted with boldface. 
   The Knowledge category means these model need external knowledge as input, while other models only need image and text themselves.
   }
   \label{tab:t32}
\end{table*}

\begin{table}[t]
   \centering
    \renewcommand\arraystretch{1.2}

       \begin{small}
       \begin{tabular}{l|l|cc}
       \Xhline{1pt}
        \textbf{Datasets} & \textbf{Models} & \textbf{Acc}(\%) &  \textbf{F1}(\%) \\ \Xhline{0.5pt}
\multirow{4}{*}{MVSA-Single} & BERT+ViT & 69.11 & 68.84 \\
& Our CLFA & 73.11  & \textbf{72.45} \\
\Xcline{2-4}{1pt}
& RoBERTa+ViT & 68.44  & 68.67 \\
& Our CLFA & \textbf{73.33}  & 72.01 \\
\hline
\hline
\multirow{4}{*}{MVSA-Multiple} & BERT+ViT & 68.14  & 67.39 \\
& Our CLFA & \textbf{69.73}  & \textbf{68.31} \\
\Xcline{2-4}{1pt}
& RoBERTa+ViT & 67.02  & 65.86 \\
& Our CLFA & 69.02  & 67.26 \\
\Xhline{1pt}
       \end{tabular}
       \end{small}

   \caption{Experimental results on MMSA.}
\label{tab:t43}
\end{table}

For the MMSD task, our proposed method CLFA achieves an F1 score of 83.91, 
outperforming all of the previous models without knowledge. 
Comparing CLFA with the basic model BERT+ViT 
without feature alignment, we get an F1 improvement with 4.11 points, 
showing that our alignment 
strategy can enhance the consistency of different embeddings.
Besides, our model is higher than InCross 
and is competitive with HKE and
CMGCN.  
The three models leverage dependency tree, sentiment of words, and object detection result of images as external knowledge.
These comparisons show that our method can 
improve the understanding ability of the model as the external knowledge can.

Among other models, multi-modal models yield better results than text-based and image-based individual models, suggesting that utilizing only uni-modal information is insufficient for the MMSD task. 
Different modalities 
complement each other, and the model needs to learn both modalities and their inter-modal relationships
to obtain better semantic understanding. Among uni-modal models, text-based models obviously outperform image-based models, 
which indicates that the text modality is much more crucial for sarcasm detection as it contains rich semantic information. 
From the user's perspective, expressing emotions and attitudes in text is more preferred, and the image serves as supplementary information to the text. 
Therefore, in multi-modal sarcasm detection, the model should focus more on understanding the text modality and utilize the auxiliary information provided by the image modality to improve the detection accuracy. 
That's also the reason why our model uses the text modality to guide the image modality in the cross attention.

For the MMSA task shown in Table \ref{tab:t43} , our method CLFA achieves an F1 score of 72.45 and 68.31 on MVSA-Single and MVSA-Multiple respectively, surpassing the baseline model by 3.61 and 0.92 points, showing that our alignment method is also effective in the multi-modal sentiment analysis task. On both datasets, CLFA brings obvious performance gains over both the BERT+ViT encoders and RoBERTa+ViT encoders, verifying that our method is effective on different backbone models. 

\subsection{Hyper-parameter Setting}
\begin{figure}
    \centering
    \includegraphics[width=0.50\textwidth]{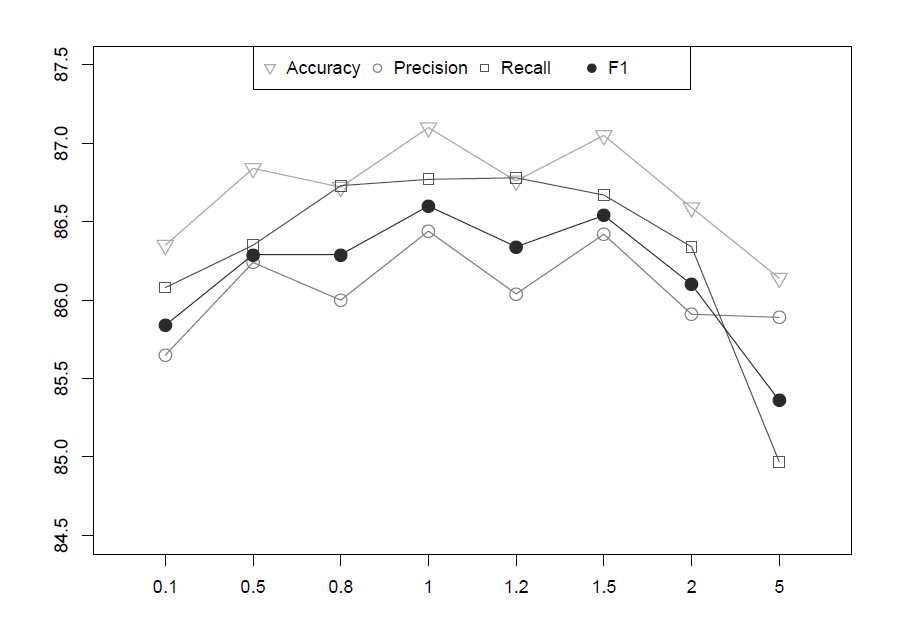}
    \caption{Experiments about $\alpha$ on the dev data of MMSD. Precision, Recall, F1 are macro average values.}
    \label{fig:para}
\end{figure}

We conduct experiments on the development set 
to investigate the coefficient $\alpha$ in Equation \ref{e47}, as shown in Figure \ref{fig:para}. When $\alpha$ is 1, we get the best performance. A larger coefficient of the contrastive learning loss function does not bring further improvement to the model. The alignment of multimodal features can provide assistance for semantic understanding,  
but excessive pursuit of feature alignment may damage the understanding of text and image by BERT and ViT pre-trained models, leading to a decrease in classification performance.

\begin{figure}[htb]
  \centering
  \subfloat[Baseline]{
    \label{sfig:base-heat}
    \includegraphics[width=0.23\textwidth]{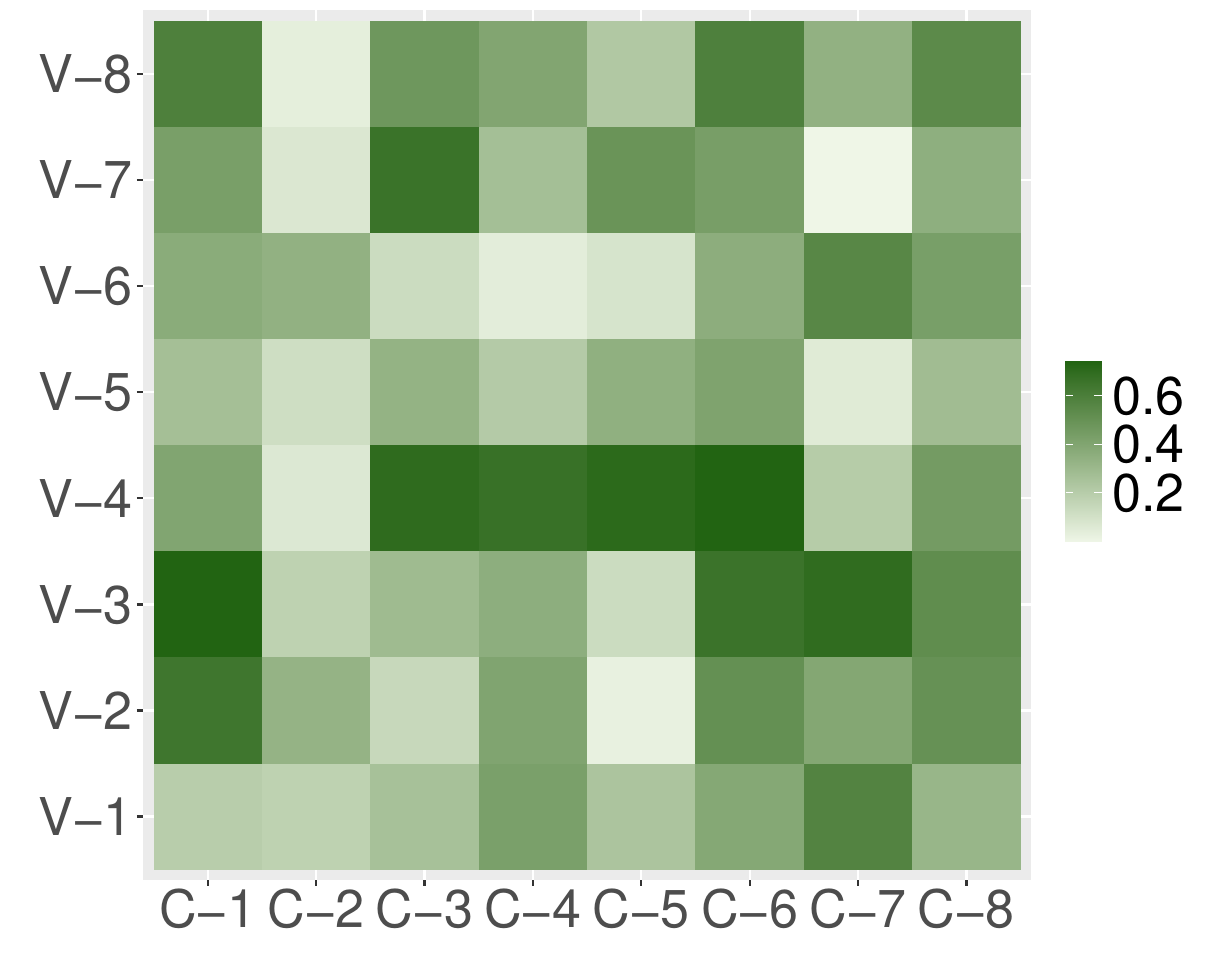}}
 \hfill
  \subfloat[CLFA]{
    \label{sfig:con-heat}
    \includegraphics[width=0.23\textwidth]{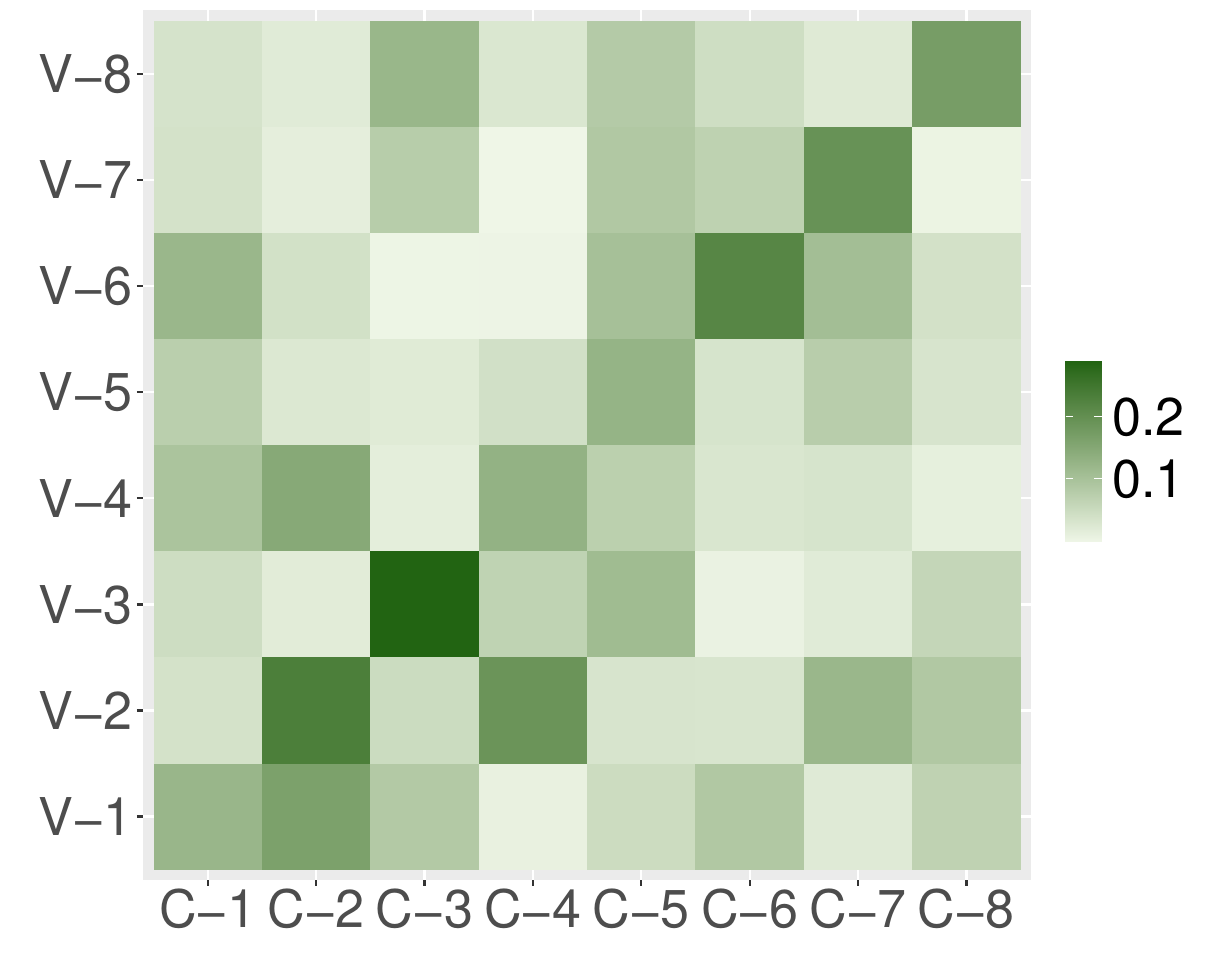}}
  \caption{Text-image alignment heatmap on the MMSD data. C indicates the image caption, V indicates the image. The darker is the color, the higher is the similarity. }
  \label{fig:heatmap}
\end{figure}

\subsection{Visual Analysis on Feature Alignment}

For MMSD and MMSA tasks, the input text and image might have different meanings, so we can't directly show the alignment effectiveness on them. But in most cases, the caption of an image has the similar meaning of it. In our experiment, we generate the image captions (textual expressions) using Clipcap \cite{mokady2021clipcap}, and leverage caption-image pairs to show the effectiveness of 
inter-modal alignment. We don't use the caption in training, so if the model can embedding the caption and image into vectors with high similarity, it can 
perform alignment well.

We present a visual analysis
in Figure \ref{fig:heatmap}. The sample images are from a mini-batch in the test dataset of MMSD. The figure shows the similarity between text modal features and image modal features, with the paired caption and image on the diagonal. We obtain the similarity value by calculating the cosine distance between the caption and image representations. 
In the baseline model, 
there shows no obvious 
mapping relationship between the image-text pairs. In contrast, in our model, the colors of the diagonal elements are 
obviously darker than those on the non-diagonal elements, indicating that the image and textual expressions  
obtain a good alignment with our CLFA method.

\subsection{Comparison with Caption Alignment}

It should be mentioned that our model CLFA uses CLIP as teacher of encoder, and achieves alignment considering CLIP as a "bridge". To show the necessary of our work, we designed another alignment method:

First, we generate the caption of image by ClipCap~\cite{mokady2021clipcap}. Second, we directly apply contrastive learning to align the embedding of the caption and the image. 

The comparison of CLFA and this method ("Caption" in the table) can be seen in Table \ref{ca}. 
CLFA obtains a 2.15 points F1 improvement above Caption alignment, showing that CLIP contains much alignment knowledge and can be a teacher in multi-modal tasks. 

\begin{table}[h]
   \centering
   \renewcommand\arraystretch{1.2}

       \begin{small}
       \begin{tabular}{l|cccc}
       \Xhline{1pt}
        \textbf{Align Methods} & \textbf{Acc}(\%) & \textbf{P}(\%) & \textbf{R}(\%) & \textbf{F1}(\%) \\ \Xhline{0.5pt}
No Align & 83.73 & 78.12 & 81.54 & 79.80 \\
Caption & 84.85 & 79.31 & 84.36 & 81.76 \\
CLFA & \textbf{86.80} & \textbf{81.51} & \textbf{86.44 }& \textbf{83.91} \\
\Xhline{1pt}
       \end{tabular}
       \end{small}

  \caption{Experimental results with different align methods.}
\label{ca}
\end{table}

\section{Further Experiments with Different Model Settings}
\subsection{Applying CLFA with Different Aggregating Methods}

\begin{table*}[t]
   \centering
   \renewcommand\arraystretch{1.2}
   \setlength\tabcolsep{7pt}
   \begin{small}
   \begin{tabular}{@{}l|cccc|ccc@{}}
   \Xhline{1pt}
   \multirow{2}{*}{\textbf{Models}} & \multirow{2}{*}{\textbf{Acc}(\%)} & \multirow{2}{*}{\textbf{P}(\%)} & \multirow{2}{*}{\textbf{R}(\%)} & \multirow{2}{*}{\textbf{F1}(\%)} & \multicolumn{3}{c}{\textbf{Macro}}  \\ \Xcline{6-8}{0.5pt} 
    & & & & & \textbf{P}(\%) & \textbf{R}(\%) &  \textbf{F1} (\%)  \\ \Xhline{0.5pt}
    Concat & 82.23 & 75.12 & 82.79 & 78.77 & 81.46 & 82.33 & 81.75 \\
    Concat+CLFA & 84.56 & 80.29 & 81.13 & 80.71 & 83.86 & 83.98 & 83.92 \\
    \Xcline{1-8}{0.5pt} 
    Co-Attention & 83.56 & 77.95 & 81.86 & 79.86 & 82.77 & 83.27 & 82.99 \\
    Co-Attention+CLFA & 84.39 & 78.76 & 83.21 & 80.93 & 83.62 & 84.19 & 83.86 \\
    \Xcline{1-8}{0.5pt} 
    Cross Attention & 83.73 & 78.12 & 81.54 & 79.80 & 82.94 & 83.41 & 83.15 \\
    Cross Attention+CLFA & \textbf{86.80} & \textbf{81.51} & \textbf{86.44} & \textbf{83.91} &\textbf{ 86.09} & \textbf{86.74} & \textbf{86.36} \\

    \Xhline{1pt}
   \end{tabular}\\[6pt]
   \end{small}

   \caption{Experimental results on MMSD with different multi-modal aggregating methods.}
   \label{agg}
\end{table*}

To 
investigate whether our CLFA model works on varying model settings, besides cross attention in our main model,  
we exploit another two multi-modal aggregating methods for further experiments. 

\textbf{Concat} It concatenates the two representations derived form text and image modalities.


\textbf{Co-Attention}~\cite{lu2016hierarchical} It treats intra-modal and inter-modal attention as different tasks. First, calculating a matrix $C$ as the similarity between text and image modalities:
\begin{equation}
\begin{split}
    C = \mathrm{tanh}(F'_tW_cF'_i)
\end{split}
\end{equation}

Then, calculating the output with $C$:

\begin{equation}
\begin{split}
    h_i&=\mathrm{tanh}(W_iF'_i+(W_tF'_t)C)\\
    h_t&=\mathrm{tanh}(W_tF'_t+(W_iF'_i)C^T)\\
    h &= Concat(h_t, h_i)\\
\end{split}
\end{equation}
where $W_c, W_t, W_i$ are trainable parameters. 

The Co-Attention method uses $C$ and $C^T$ to measure the relationship between two modalities, and keep the intra-modal information as a linear layer.

\textbf{Results} The experimental results are reported in Table \ref{agg}. 
To our expectation, CLFA works well on all of three aggregating methods, 
obtaining 1.94, 1.07 and 4.11 points F1 improvement on Concat, Co-Attention and Cross-Attention, respectively. The improvement on cross attention is highest, because a well-aligned text-image representation enables cross attention 
to capture cross-modal relationship. 
That is, if the input data is well aligned, the cross-attention works better. Besides, the main difficulty in MMSD and MMSA tasks is capturing the inconsistency between modalities, so cross attention with feature alignment achieves better performance.

\subsection{Applying CLFA to the Knowledge-enhanced Model}

\begin{table*}[t]
   \centering
   \renewcommand\arraystretch{1.2}
   \setlength\tabcolsep{7pt}
   \begin{small}
   \begin{tabular}{@{}l|cccc|ccc@{}}
   \Xhline{1pt}
   \multirow{2}{*}{\textbf{Models}} & \multirow{2}{*}{\textbf{Acc}(\%)} & \multirow{2}{*}{\textbf{P}(\%)} & \multirow{2}{*}{\textbf{R}(\%)} & \multirow{2}{*}{\textbf{F1}(\%)} & \multicolumn{3}{c}{\textbf{Macro}}  \\ \Xcline{6-8}{0.5pt} 
    & & & & & \textbf{P}(\%) & \textbf{R}(\%) &  \textbf{F1} (\%)  \\ \Xhline{0.5pt}
    ViT+BERT (Cross Attention) & 83.73 & 78.12 & 81.54 & 79.80 & 82.94 & 83.41 & 83.15 \\
    ~~~~~~~~~+Knowledge & 86.38 & 81.15 & 85.71 & 83.37 & 85.67 & 86.27 & 85.92 \\
    ~~~~~~~~~+Knowledge+CLFA & \textbf{87.30} & \textbf{81.73} & \textbf{87.70} & \textbf{84.61} & \textbf{86.59} & \textbf{87.36} & \textbf{86.90} \\ 
    \Xhline{1pt}
   \end{tabular}\\[6pt]
   \end{small}

   \caption{Experimental results on MMSD with knowledge-enhanced methods.}
   \label{ke}
\end{table*}

As listed in the Section Baseline Methods, many recent works introduce knowledge to enhance the model's ability for multi-modal semantic understanding. Actually, our model CLFA for feature alignment is orthogonal to the knowledge methods, and can also take external knowledge as input and get higher performance. 

To prove this assumption, 
we design a knowledge-enhanced CLFA, 
integrating OCR results and word sentiment in the cross-attention layers. Concretely, we modify the three cross-attention layers as below: 

\textbf{First layer} In this layer, we extract sentiment knowledge from SenticNet~\cite{cambria-etal-2022-senticnet} to better understanding the text data. So we change this layer to do self-attention on the text only:

\begin{equation}
\begin{split}
    h_1=Attn'(W_qF'_t, W_kF'_t, W_vF'_t)
\end{split}
\end{equation}

For every word $x$ in SenticNet, the data contains a value $S_x$ in $[-1, 1]$ to represent the sentiment of it. In our model, we adopt the same way as CMGCN~\cite{liang-etal-2022-multi}:

\begin{equation}
\begin{split}
    SC(x, y)&=|S_x - S_y| e^{-S_xS_y}\\
    Attn'(Q,K,V)&=\mathrm{softmax}(\frac{(QK^T) \times (1 + SC)}{\sqrt{d_k}})V\\
\end{split}
\end{equation}

\textbf{Second layer} This layer performs text-image cross-attention:

\begin{equation}
\begin{split}
    h_2=Attn(W_qh_1, W_kF'_i, W_vF'_i)
\end{split}
\end{equation}
Since SenticNet can't deal with the image, we use normal attention here.

\textbf{Third layer} We use a third-party OCR tool to extract the expression $F'_{OCR}$ and then compute attention:

\begin{equation}
\begin{split}
    h_3=Attn'(W_qh_2, W_kF'_{OCR}, W_vF'_{OCR})
\end{split}
\end{equation}

\textbf{Results} The results of knowledge experiments are reported in Table \ref{ke}. The external knowledge really improves performance by a large margin, with 3.57 points improvements in F1 score. Even  
based on the strong knowledge-enhanced model, our method CLFA
achieves better results on all metrics, obtaining 1.24 F1 points  improvement. It demonstrates the powerful ability of CLFA for multi-modal semantic understanding.  

\section{Related Work}

\subsection{Multi-modal Sarcasm Detection}

In MMSD task, many neural network models have been proposed. \citet{cai-etal-2019-multi} propose a hierarchical fusion model to integrate different modalities. \citet{xu-etal-2020-reasoning} propose a Decomposition and Relation Network.
that simultaneously learns the relative and relational information between multiple modalities. 
\citet{pan-etal-2020-modeling} emphasize the importance of identifying the inconsistencies within and between modalities. 
\citet{liang2021multi} propose an interactive graph convolutional neural network to learn the emotional inconsistencies within and between modalities. 
\citet{liu-etal-2022-towards-multi-modal} 
claim that the inconsistencies of sarcasm exist not only at the atomic level of individual words but also at more complex compositional levels. 
\citet{liang-etal-2022-multi} 
use FastRCNN~\cite{girshick2015fast} to identify objects and their attributes in images and then leverage external knowledge to construct an inter-modal graph.


\subsection{Multi-modal Sentiment Analysis}


In MMSA task, CoMN~\cite{xu2018co} uses memory network to do cross-modal fusion, and Self-MM~\cite{yu2021learning} uses joint training on both unimodal task and cross-modal task, which can both be designed with MVSA dataset. \citet{xu2023borrowing} presents a method using teacher-student framework and extracting features from user comments.

\subsection{Multi-modal Alignment}
Multi-modal learning aims to overcome the limitations of individual modalities by leveraging the complementary advantages of different modalities, so alignment is an important task in this area.
Its advantage lies in the ability to reduce noise and ambiguity from a single modality by combining multiple modalities, enabling the model to have a more comprehensive and accurate understanding of the input data. 

The goal of multi-modal alignment is 
to learn the correspondence between two or more modalities. 
Currently, there are two main approaches for multi-modal alignment: attention mechanisms and contrastive learning. 
\citet{yue2018attentional} employ attention to align different regions in images and text. 
ViLBERT~\cite{lu2019vilbert} utilizes a Transformer structure with co-attention to align image-text pairs. 

On the other hand, contrastive learning has achieved success in coarse-grained multi-modal 
alignment. Representative works include CLIP~\cite{radford2021learning} and ALIGN~\cite{jia2021scaling}. 

Feature fusion is widely applied for multi-modal learning. 
The simple method is concatenating or addition of different features to obtain the overall representation.  
There are also methods based on matrix operations, such as  Multi-modal Compact Bilinear (MCB)~\cite{fukui-etal-2016-multimodal}, Multi-modal Low-rank Bilinear (MLB)~\cite{kim2016hadamard}, Multi-modal Factorized Bilinear Pooling (MFB)~\cite{yu2017multi}. What's more, attention-based feature fusion is widely used in VQA tasks, such as co-attention~\cite{yang-etal-2022-low}
and relational attention~\cite{wu2018object}.



\section{Conclusion}

In this paper, we propose a cross-modal feature alignment approach called CLFA. We introduce CLIP as a teacher model 
in learning multi-modal feature alignment, 
which enables the model to effectively perform 
cross-modal interaction 
during feature fusion. Experiment results show that CLFA gains large improvement on MMSA and MMSD tasks, and can be combined with other aggregating methods and knowledge-enhanced models. Through visual analysis, we can see that CLFA can align text and image well. In conclusion, CLFA is an effective and flexible model which can be used in cross-modal semantic understanding tasks.

\section*{Acknowledgement}
This work is supported by the National Natural Science Foundation of China (62076008) and the Key Project of Natural Science Foundation of China (61936012).
\section*{Bibliographical References}\label{sec:reference}

\bibliographystyle{lrec-coling2024-natbib}
\bibliography{lrec-coling2024-example}


\end{document}